# Detecting The Objects on The Road Using Modular Lightweight Network

*Abstract*—This paper presents a modular lightweight network model for road objects detection, such as car, pedestrian and cyclist, especially when they are far away from the camera and their sizes are small. Great advances have been made for the deep networks, but small objects detection is still a challenging task. In order to solve this problem, majority of existing methods utilize complicated network or bigger image size, which generally leads to higher computation cost. The proposed network model is referred to as modular feature fusion detector (MFFD), using a fast and efficient network architecture for detecting small objects. The contribution lies in the following aspects: 1) Two base modules have been designed for efficient computation: Front module reduce the information loss from raw input images; Tinier module decrease model size and computation cost, while ensuring the detection accuracy. 2) By stacking the base modules, we design a context features fusion framework for multi-scale object detection. 3) The propose method is efficient in terms of model size and computation cost, which is applicable for resource limited devices, such as embedded systems for advanced driver assistance systems (ADAS). Comparisons with the state-of-the-arts on the challenging KITTI dataset reveal the superiority of the proposed method. Especially, 100 fps can be achieved on the embedded GPUs such as Jetson TX2.

## I. INTRODUCTION

Convolutional Neural Networks (CNNs) have produced impressive performance improvements in many computer vision tasks, such as image classification [1-7], object detection [8-16], and image segmentation [17-20]. Since AlexNet [5] popularized deep convolutional neural networks by winning the ImageNet Challenge ILSVRC 2012 [21], many innovative CNNs network structures have been proposed. Szegedy et al. [4] propose an "Inception" module which is comprised of a number of different sized filters. He et al. [1] propose ResNet with skip connection over multiple layers. Huang et al. [2] propose DenseNet with direct connection from any layer to all subsequent layers, which enables training very deep networks with more than 200 layers. Due to these excellent network structures, the quality of many vision tasks has been growing at a dramatic pace. Among them, object detection is one of the most benefited areas because of its wide applications in intelligent monitoring, security system, and autonomous driving.

State-of-the-art CNNs based object detection methods have been proposed in the past several years, such as RCNN [22], Faster RCNN [12], YOLO [23], SSD [10], etc. These detection system has greatly improved the accuracy of detection tasks. However, these detection methods have two notable limitations:

1) Network architectures are deep and complicate. In order to achieve higher accuracy, the general trend is to make deeper and more complicate networks. Recent evidence reveals that network depth is of crucial importance, and the leading results on the challenging ImageNet dataset all exploit "very deep" models. Such as VGG [4] and ResNet [24], with layers from 19 to more than 100, both won the 1st place on the ImageNet ILSVRC classification task.

Many nontrivial object detection tasks have also greatly benefited from very deep models. For example, Faster RCNN and SSD both use VGG as backbone network. However, these advances to improve accuracy are not necessarily making networks more efficient with respect to size and speed. One of the problems brought about by the continuous deepening network structure is that the model size is getting larger, and the budget of computation is getting higher. In many real world applications such as robotics, autonomous driving and augmented reality, the object detection tasks need to be carried out in a timely fashion on a computationally limited platform.

2) Poor performance in detecting small objects. Most of early object detection methods use only the last feature layer for detection, for example, R-CNN uses the region proposals from selective search [25] or Edge boxes [26] to generate the region-based feature and SVMs are adopted to do classification. YOLO [27] divides the input image into several grids and performs localization and classification on each part using the last feature map layer. Generally, last layer corresponds to bigger receptive field in the original image, which makes the small objects detection became difficult.

Moreover, these object proposals methods have been proven to achieve high recall performance and satisfactory detection accuracy in the popular ILSVRC [21] and PASCAL VOC [28] detection benchmark, which require loose criteria, i.e. a detection is regarded as correct if the intersection over union (IoU) overlap is more than 0.5. However, these object proposal methods fail under strict criteria (e.g. IoU> 0.7) for example in the considerable challenge KITTI [29] benchmark, their performance is barely satisfactory.

Recent advances of FCN[30], SSD [10] and YOLOv3[31] cast a new light on the fine granular analysis by fusing the detection of multiscale feature maps, which also inspire the model design of this paper.

In addressing the above issues, this paper's model design principle is: pursuing the competitive accuracy for small objects in higher speed, which may be used for computationally and memory limited platforms such as drones, robots, and phones. We develop a novel detector called modular feature fusion detector (MFFD). The contributions of this paper are summarized as follows:

1) Two simple yet effective base modules have been designed as the building blocks of the network: Front module reduce the information loss from raw input images by utilizing more convolution layers with small size filters; Tinier module use pointwise convolution layers before conventional convolution layer to decrease model size and computation, while ensuring the detection accuracy.

2) By combing the building blocks, modules, in different ways, we can generate modular detection framework efficiently which is capable of fusing the contextual information from different scales for small object detection.

3) The proposed model is lightweight in terms model size and computation cost, therefore it is applicable for the embedded system, such as ADAS. Evaluations on a widely used KITTI dataset demonstrate the effectiveness of the presented detection method in terms of accuracy and speed.

The rest of the paper is organized as follows. Section 2 provides a brief summarization of the related works; Section 3 detail the Front module, Tinier module, and describe MFFD architecture. Experimental results and comparisons are provided in Section 4. Finally, we conclude this work in Section 5.

## II. RELATED WORK

In this section, we are going to briefly review the advances of the related works from three aspects: object detection framework, the lightweight network design and the small object detection.

### A. Object detect framework

In the past few years, mainly due to the advances of deep learning, more specifically convolutional neural networks [1, 2, 4, 5], the performance of object detection has been improved at a dramatic pace. RCNN [9] is the first to show that a CNNs can lead to dramatically higher object detection performance on PASCAL VOC. RCNN contains four steps: 1) Using selective search [25] to extracts around 2000 bottom-up region proposals from an input image; 2) Computing features for each proposal using a convolutional neural networks (CNNs); 3) Classifying each potential object region using class-specific linear SVMs; 4) Post-processing is used to refine the bounding boxes, eliminate duplicate detections, and rescore the boxes based on other objects in the scene. R-CNN requires high computational costs since each region is processed by the CNNs network separately. Fast R-CNN [32] and Faster R-CNN [12] improve the efficiency by sharing computation and using neural networks to generate the region proposals.

RCNN, Fast RCNN, Faster RCNN and other variants achieve excellent object detection accuracy by using a deep CNNs to classify object proposals. They can be summarized as region proposal based methods. However, these region-proposal based methods have several notable drawbacks: 1) Training is a multi-stage pipeline; 2) Training is expensive in space and time; 3) Object detection is slow. Furthermore, some real-time object detection methods are proposed, such as YOLO [11], YOLOv2 [33] and SSD [10]. These methods can be summarized as regression based methods. YOLO uses a single feed-forward convolutional network to directly predict object classes and locations, which can achieve a good tradeoff between speed and accuracy.

SSD [10] discretizes the output space of bounding boxes into a set of default boxes over different aspect ratios and scales. It improves YOLO[11] in several aspects: 1) Using small convolutional filters to predict categories and anchor offsets for bounding box locations; 2) Using pyramid features for prediction at different scales; 3) Using default boxes and aspect ratios for adjusting varying object shapes. Regression based methods obtains competitive accuracy and breaks the speed bottleneck.

### B. Lightweight network design

For CNN models, network architecture designs play an important role. In order to achieve higher accuracy, building deeper and larger convolutional neural networks (CNNs) is a primary trend. Recent evidence [3, 24, 34, 35] also reveals that network depth is of crucial importance, and the breakthroughs on various benchmark datasets all exploit "very deep" models. Larger networks perform well on GPU-based machines, however, deeper network architectures are often unsuitable for smaller devices with limited memory and computation power.

Recently, the increasing demands of running high quality deep neural networks on computationally and memory limited devices encourage the study on lightweight model designs. For example, Han et al. [36] proposed "deep compression" with three-stage pipelines: pruning, quantization and entropy coding, which can reduce the storage requirement of neural networks. Forrest et al. [37] design Fire module to build SqueezeNet, which reduces parameters and computation significantly while maintaining accuracy.

Andrew et al. [37] propose MobileNet that uses depth wise separable convolutions to build lightweight networks. Zhan et al. [38] propose ShuffleNet which utilizes two operations, such as pointwise group convolution and channel shuffle. Mark et al. [39] present a new mobile architecture named MobileNetv2 which is based on an inverted residual structure. This work significantly decreasing the number of operations and memory cost while retaining the same accuracy. Prior works indicate that lightweight network architecture can also achieve great performance with less computation and time cost.

*C. Small objects detection*

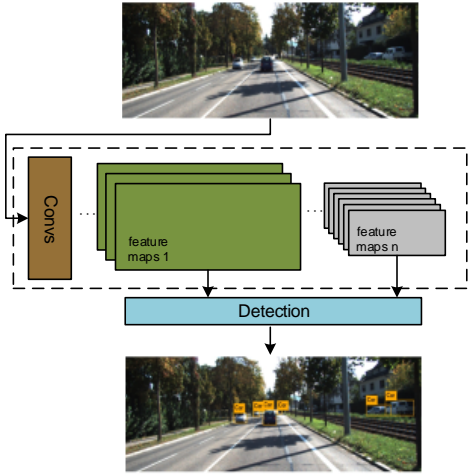

Fig. 1. The pipe line of small object detection by combing the output of multiple feature maps.

Though tremendous advances have been made in object detection, one of the remaining open challenges is detecting small objects. Context information is the key to finding small objects. Bell et al. [40] introduce top-down structure to combine the different level features together to enhance the ability for small object detection. For pedestrian detection, Park et al. [41] uses ground plane estimation as contextual features and improve detection for small instances. For face detection, Zhu et al. [42] simultaneously pool ROI features around faces and bodies to score the detections and show improvements on small object detection. Recently, SSD [10] and YOLOv3[31] cast a new light on the fine granular analysis by fusing the detection of multiscale feature maps, as shown in Fig. 1.

RRC [43] is one of the recent breakthroughs for small object detection, which based on the recurrent rolling convolution architecture over multi-scale feature maps to construct object classifiers and bounding box regressors. Evaluations on KITTI dataset proves that contextual information is important for improving the small objects detection. However, there are wo shortcomings for RRC: 1) The training time is long. The recurrent rolling convolution architecture improve the detection performance, but this also leads to longer training time. 2) High resolution input image leads to high computation cost. RRC extends image size by factor of 2 in the input layer, which will increase the computation by a big margin. These shortcomings make RRC not directly applicable for smaller devices with limited memory and computation power.

Thanks to these excellent object detection and feature fusion methods, they inspire the network design in this work. The MFFD is built upon the YOLO framework, and it frame object detection as a regression problem to spatially separated bounding boxes and associated class probabilities.

## III. MODULAR LIGHTWEIGHT NETWORK

Convolution neural networks (CNNs) have shown significant improvements in computer vision tasks [1-3, 5, 10, 17, 32, 44], in which network architecture designs play an important role. In order to achieve higher accuracy, building deeper and larger convolutional neural networks is a primary trend, such as [2, 3, 45, 46]. However, deeper network architectures are often unsuitable for smaller devices with limited memory and computation power. Therefore, we have developed a modular lightweight network model which was especially designed for the computationally and memory limited devices, such as advanced driver assistance systems (ADAS).

The proposed network is referred to as modular feature fusion detector (MFFD). It contains two major modules for efficient computation: Front module reduce the information loss from raw input images by utilizing more convolution layers with small size filters; Tinier module use pointwise convolution layers before conventional convolution layer to decrease model size and computation, while ensuring the detection accuracy. By stacking these modules into the detection pipeline, a lightweight network that capable for small object detection can be developed.

*A. Network parameter reduction*

It is known that the number of parameters in a convolutional layer is related to the number of input channels, the size of filters, and the number of filers. Consider a convolution layer that is comprised entirely of 3×3 filters, the total quantity of parameters in this layer can be computed as:

$$T_l = M_l \times N_l \times 3 \times 3 \qquad (2)$$

where $T_l$ is the total quantity of parameters in the $l^{th}$ layer, $M_l$ stands for the channels of input feature map, $N_l$ is the number of filters, and also denotes the number of output channels.

Therefore, there are two strategies to reduce the number of parameters and computational complexity: (1) decrease the number of 3×3 filters, which is referred to as $N_l$; (2) decrease the number of input channels of the 3x3 convolution layer, which is referred to as $M_l$. We choose the second strategy, and reduce the parameters by using a 1×1 convolution (also called pointwise convolution in [37]) before each 3×3 convolution. Fig.2 shows the structure difference between the traditional 3×3 convolution and the updated one used in the proposed method.

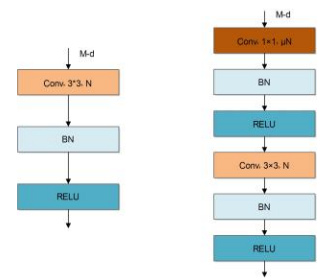

Fig. 2. The structure change of the traditional 3×3 convolution layer. Left, conventional 3×3 convolution layer with batch normalization and ReLU; right, the convolution layer used in this work.

The 1×1 convolution in our method have two important effects:

(1) Adding non-linearity. In conventional CNNs, convolution layers take inner product of the linear filter and the underlying receptive field followed by a nonlinear activation function at every local portion of the input. This linear convolution is sufficient for abstraction when the instances of the latent concepts are linearly separable. However, the features we wish to extract are highly non-linear in general which leads to an over-complete set of filters to cover the variations. The 1×1 convolution before the conventional 3×3 convolution corresponds to a linear combination of different channels for each pixel. Since the input of the convolution layer is a three-dimensional structure with multiple channels instead of a planar structure, the network model can greatly achieve the spatial information fusion of the feature maps from different channels. This operation is efficient to increase the non-linearity of the decision function without affecting the receptive fields and the representation power of neural networks.

(2) Decreasing the number of parameters. In the deep convolution neural network, there is redundancy in the output feature maps [47, 48]. The method used in [47] is to design a threshold function to evaluate the importance of output feature maps, and pruning unimportant channels. We use another method that decreasing the input dimensions through the 1×1 convolution, not only to reduce the model parameters, but also to reduce the impact of the redundancy of the feature map.

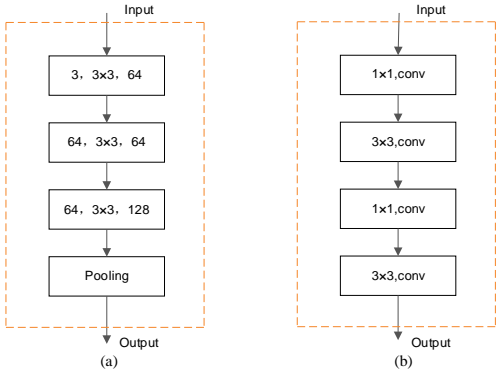

Fig. 3. The structure of Front module (a) and Tinier module (b).

### B. Building blocks: Tinier module and Front module

With the trend of designing very deep CNNs, state-of-the-art networks [34, 37, 44] usually consist of repeated building blocks or modules with the same structure. Generally, carefully designed blocks are able to achieve better performance with low complexity. So, in order to take advantages of module structure, we proposed the Tinier module and Front module. As shown in Fig.3 (b), a Tinier module is comprised of two 1×1 convolution layers and two 3×3 convolution layers, where the 1×1 layer is for dimension reduction. Each convolution layer is followed by batch normalization and ReLU [49] operations.

Inspired by Inception-v3 [44] and v4 [34], we define Front module as a stack of three 3×3 convolution layers followed by a 2×2 max pooling layer as shown in Fig.3(a). The first convolution layer works with stride = 2 and the other two are with stride = 1. We observe that adding this simple Front module can evidently improve the model performance in our experiments. A possible explanation is that, since the downsmapling is applied at the end of front module, convolution layers have large feature maps which may reduce the information loss from raw input images.

### C. Modular feature fusion detector (MFFD)

Based on the building blocks: Tinier module and Front module, proposed in previous section, we are going to present a multi-scale object detection framework. The basic idea is also fusing the information from the feature maps with different context information, as shown in Fig.1. The difference is that the network is not build from individual layers, but from available modules which is designed for efficient computation. That's the reason why the proposed method is called modular detector.

In order to make a straightforward comparison, we are going to present three models. The first model is a single track detector, similar to YOLOv2, and there is no feature fusion in the network. This model is used as the reference model. In addition, two feature fusion models are presented using different strategies. We are going to detail these models in the following parts.

The reference single-track detection model is shown in Fig.4(a). Our target is to build a lightweight detector with competitive performance, therefore the reference detector is relatively simple, which contains one Front module and a stack of four Tinier modules. The final detector layer is a 1×1 convolution layer with linear activation, whose definition is similar to YOLOv2 [33]. It performs max-pooling with a stride of 2 after every Tinier module except the last one. The full reference architecture is detailed in Table 1. For detector layer, the number of filters is decided by the number of classes, therefore we denoted it as *n*. Take VOC as an example, prediction of 5 boxes with 5 coordinates each and 20 classes per box leads to 125 output filters. The total parameter number of reference model is 3.6M, less than other small models, such as Fast YOLO and Tiny YOLO, which are 27.1M and 15.8M, respectively. (It should be noted that here we calculate the number of parameters, not the model size)

TABLE I. DETAILED INFORMATION OF THE REFERENCE MODEL. $N_{1\times1}$ AND $N_{3\times3}$ REPRESENTS THE NUMBER OF 1×1 AND 3×3 FILTERS RESPECTIVELY.

| Module | Output size | Filter size/stride | $N_{1\times1}$ | $N_{3\times3}$ | Param. |
|---|---|---|---|---|---|
| Input | 320×576×3 | N/A | N/A | N/A | N/A |
| Front | 80×144×128 | $\begin{bmatrix}3\times3/2\\3\times3/1\\3\times3/1\end{bmatrix}$ | 0 | $\begin{bmatrix}64\\64\\128\end{bmatrix}$ | 112k |
| Tin.1 | 40×72×128 | $\begin{bmatrix}1\times1/1\\3\times3/1\end{bmatrix}\times2$ | 16 | 128 | 40k |
| Tin.2 | 20×36×256 | $\begin{bmatrix}1\times1/1\\3\times3/1\end{bmatrix}\times2$ | 32 | 256 | 159k |
| Tin.3 | 20×36×512 | $\begin{bmatrix}1\times1/1\\3\times3/1\end{bmatrix}\times2$ | 64 | 512 | 636k |
| Tin.4 | 10×18×1024 | $\begin{bmatrix}1\times1/1\\3\times3/1\end{bmatrix}\times2$ | 128 | 1024 | 3155k |
| Det. | 10×18×*n* | [1×1/1] | *n* | 0 | 13k,n=125 |
| | | | | | 3.6M (total) |

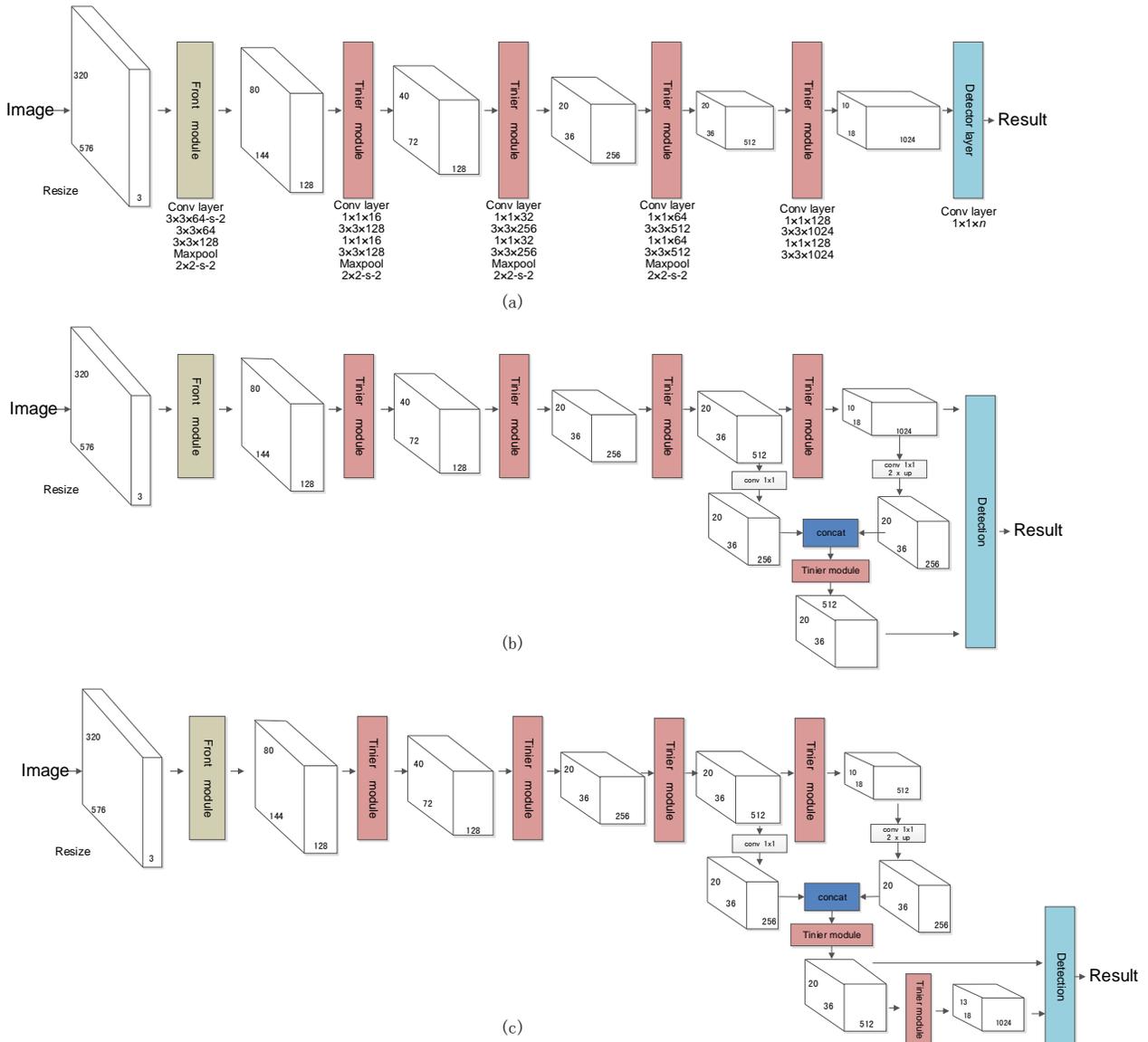

Fig. 4. The block diagrams of the modular detectors presented in this paper. (a) Single-track detector that used as the reference model. (b) Pre-context fusion model that is referred to as MFFD_A and (c) and Post-context fusion model that is referred to as MFFD_B.

Starting from the reference model, in order to utilize the context information for small object detection, we developed two models to fuse the different level features and predict objects in the fused feature maps. The structure of these two models are shown in Fig.4 (b) and (c).

The first fusion model is pre-context fusion model that is referred to as MFFD_A in Fig.4 (b). There are two tracks in this detector: 1) Low resolution track which takes exactly the same structure as the reference model in Fig.4 (a) and perform the detection on 10×18 scale; 2) High resolution track fuse the feature map of Tin.3 and up-sampled feature map of Tin.4 by concatenation, and performed the detection on 20×36 scale. High detection speed is one of our goals, but more fusion operations and large feature map fusion will increase computation cost, therefore we only fuse the features of the last two layers to get a good tradeoff between the speed and accuracy. Since the low resolution detection is applied before the feature fusion, this model is called pre-context fusion model.

The second fusion model is post-context fusion model that is referred to as MFFD_B in Fig.4 (c). In order to explore the context information for both low and high resolution detection, we postpone both resolution detection after the feature fusion. Specifically, 1) The Front module and Tin.1-Tin.4 modules are copied from the reference model (the only difference is that in Tin.4, the number of 1×1 convolution is changed from 1024 to 512 to reduce the computation). 2) The features maps from Tin.3 and Tin.4 are concatenated at 20×36 scale and the high resolution detection is performed on this scale. 3) One more Tinier module is applied to reduce the scale and low resolution detection is performed on 10×18 scale.

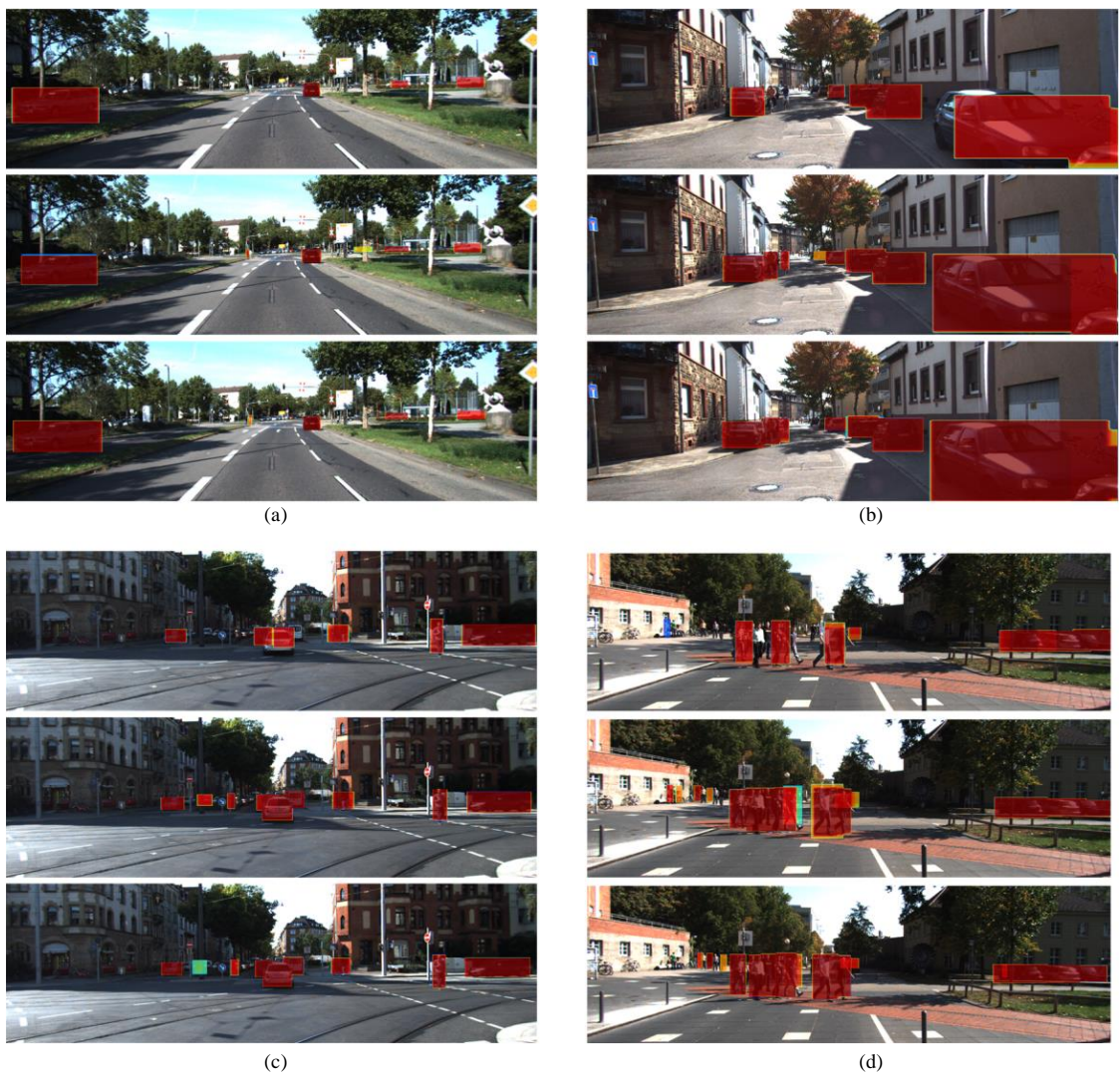

Fig. 5. The detection heat maps of all the three models. (a)-(d) provide the heat map images of 4 different street views. From top to bottom, the heat maps of reference model, MFFD_A and MFFD_B are given.

Comparing the models in Fig.4, all the three models have similar backbone network structure. MFFD_A add one more Tinier module to high resolution detection and MFFD_B add two Tinier modules to incorporate the context information for both two resolution detections. Therefore, the computation cost of MFFD_B is slightly higher than the others.

In order to have an intuitive understanding of the contribution of the feature fusion, we plot the detection heat maps of all the three models, as shown in Fig.5. Four street views with the objects of different scales are shown in (a)–(c) and the heat maps of reference model, MFFD_A and MFFD_B are listed from top to bottom. It is clear that contribution of feature fusion is very impressive, and the detection has been improved in two aspects: firstly, by using the context information the localization of the MFFD models are more accurate than the reference model; secondly, the small objects, such as pedestrian and the cars that far from the camera, show strong response for the MFFD models.

IV. EXPERIMENTS

In this section, we evaluate the proposed MFFD on the KITTI detection benchmark dataset. First, we investigate some important design factors about the MFFD method. Then, we compare proposed MFFD with the state-of-the-art lightweight models. Finally, we analyze the performance of different models in terms of precision, model size and inference time on the different platforms, such as GPU, CPU and some embedded GPUs.

*A. Dataset and training details*

**Dataset**. Since the background of this research is the object detection for ADAS, the detection targets are mainly the objects on the road, such as pedestrian or vehicles. Therefore, the benchmarking dataset used in this section is the well-defined KITTI [50] detection dataset. The KITTI dataset consists of 7481 images for training and 7581 images for testing, and a total 80256 labeled objects. Specifically, in this paper, most widely used three categories: Car, Pedestrian, and Cyclist, are used for evaluation.

Evaluation for each class has three difficulty levels: Easy, Moderate, and Hard, which are defined in term of the occlusion, size and truncation levels of objects. Checking [50] for detailed definition of these difficulty levels. Since the ground truth labels of the test set are not publicly available for researchers, following [51], we partition the KITTI training images set into training and validations sets to evaluate our approach.

It should be noted that training individual detectors for different categories with different parameter settings, such as input image sizes, can possibly boosting the performance on the benchmarking. For our case, since the application is ADAS, not only the accuracy but also the computation cost need to be considered. Therefore, all the evaluations are provided with a single detector with multi-category detection outputs.

**Implementation**: The backend deep learning framework is Darknet introduced by Redmon and Farhadi [27, 31], which is a lightweight framework written in C and can be easily deployed in many platforms.

The image number of KITTI dataset is not enough for deep model training and model pretrain is necessary for the model parameter initialization. We pretrain the MFFD on the ILSVRC 2012 classification dataset for 16 epochs using stochastic gradient descent (SGD) with a starting learning rate of 0.1. Based on the pretrained model, we then train the MFFD for detection on KITTI dataset on Nvidia TianX GPU with a batch size of 4. This batch size requires a small amount of GPU memory so the proposed model can be easily trained within a limited time. For learning, stochastic gradient descent (SGD) with momentum of 0.9 was used for optimization. Weight decay is set to 0.0005. The MFFD is trained for total of 160 epochs with a starting learning rate 0.001, dividing it by 10 at 60 and 90 epochs. During training, we use standard data augmentation tricks including rotations, and hue, saturation, and exposure shifts.

*B. Evaluation of feature fusion*

In this subsection, we are going to evaluate the contribution of the feature fusion quantitatively. In these experiments, following [28], a predicted bounding box is correct if its intersection over union (IOU) with the ground truth is higher than 0.5 which is used in PASCAL VOC test. The difficulty level is not considered in this subsection, so the test results are slightly different from the following section. We adopt the mean average precision (mAP) as the metric for evaluation detection performance.

There four variants are evaluated. Single-scale detector, whose structure is the reference model as shown in Fig.4(a) which has single detection track and the final feature maps size is 10×18×1024. Multi-scale detector's structure is very close to the single-scale detector, and the only difference is that detection layer is connected to the outputs from both Tin.3 and Tin.4 with feature maps $20 \times 36 \times 512$ and $10 \times 18 \times 1024$ respectively. This multi-scale pipeline is inspired by the SSD [10] and the defection is applied on two separate feature maps without feature fusion. MFFD_A and MFFD_B are the feature fusion model introduced in this paper and whose structure is show in Fig.4 (b) and (c).

The evaluation results are summarized in Table 1, where the best ones are marked as bold-red and the second runners are marked as bold-blue. As shown in Table 1, if we only predict objects in the single scale with spatial feature size 10×18, the mAP (row 1) is 56.95. If we apply the detection on two difference spatial scales, $10 \times 18$ and $20 \times 36$, the mAP is increased to 61.13 (row 2). In particular, the performance of small objects detection such as Pedestrian and Cyclist is improved significantly. Which proves that larger feature maps benefit to small objects detection.

In addition, besides the multiscale detection, the feature fusion modules are applied in MFFD_A (row 3) and MFFD_B (row 4), and more improvement can be observed. They beat the multi-scale detector by a big margin, and increase the mAP by 7.35% and 11.58%. These promising results benefit from the use of contextual information. Moreover, the results show that MFFD_B fusion module is 4.3% better than MFFD_A fusion module. It should be noted that MFFD_A fusion module is more competitive in term of speed, which will be introduced in following subsections.

Since the focus of this paper is modular feature fusion detector, the single scale and multi scale detector are not going to be discussed in more details and quantitative evaluations against the-state-of-the-arts are only applied for the MFFD models.

Table 1. Contribution of feature fusion

| Model | mAP | Car | Pedestrian | Cyclist |
|---|---|---|---|---|
| Single-scale | 56.95 | 78.93 | 47.78 | 44.13 |
| Multi-scale | 61.13 | 80.37 | 52.39 | 50.62 |
| MFFD_A | **68.48** | **88.49** | **61.27** | **55.49** |
| MFFD_B | **72.71** | **88.66** | **62.54** | **66.93** |

*C. Comparison with the-state-of-the-arts*

In this subsection, the proposed MFFD detectors are compared with the state-of-the-arts that designed for road objects detection. There are some very promising models, such as RRC[43], but since our work is a lightweight model and both accuracy and computation cost are considered, therefore we select some lightweight baselines for comparison. The selected lightweight baselines are: Pose-RCNN[52], Faster-RCNN[12], FYSqueeze[53], HNet[54], tiny-det[55], ReSqueeze [56], Vote3Deep [57]. For evaluation, we compute precision-recall curves for car, pedestrian and cyclist in different difficulty levels, which are used in official KITTI rankings. According to the standard KITTI setup, the IoU threshold is 70% for car, and 50% for pedestrian and cyclist. To rank different methods, we also compute average precision (AP).

Fig. 6 plots the precision-recall curves for all the baselines. From top to bottom, the results on different categories are given, such as car, pedestrian and cyclist. From left to right, the results of different difficulty levels are presented: easy, moderate and hard. The proposed methods MFFD_A and MFFD_B are plotted in blue curves. We can see that for most of the cases, MFFD detectors yield very competitive performance.

In order to have a more intuitive comparison for all the methods, Table 2 gives the average precision (AP) of object detection for different categories in different levels. In Table 2, we mark the best methods in bold-red and the runner-ups in bold-blue. It can be seen that the proposed MFFD_A gets the best results in the case of Car-Moderate and Pedestrian-Hard and gets the runner-up results in the case of Car-Easy and Cyclist-Hard. MFFD_B gets the best results in the case of Car-Easy and the runner-up in the case of Car-Hard, Pedestrian-Hard and Cyclist-Mod. It should be noted that although MFFD didn't get the best results in all the cases, it is clear that the gap between the MFFD and the best method is small. For example, in the case of Pedsetrian-Mod, the precision of the MFFD is 65.46, which is 1.28 slower than the best method. Fig. 7 shows some detection examples on KITTI testing set.

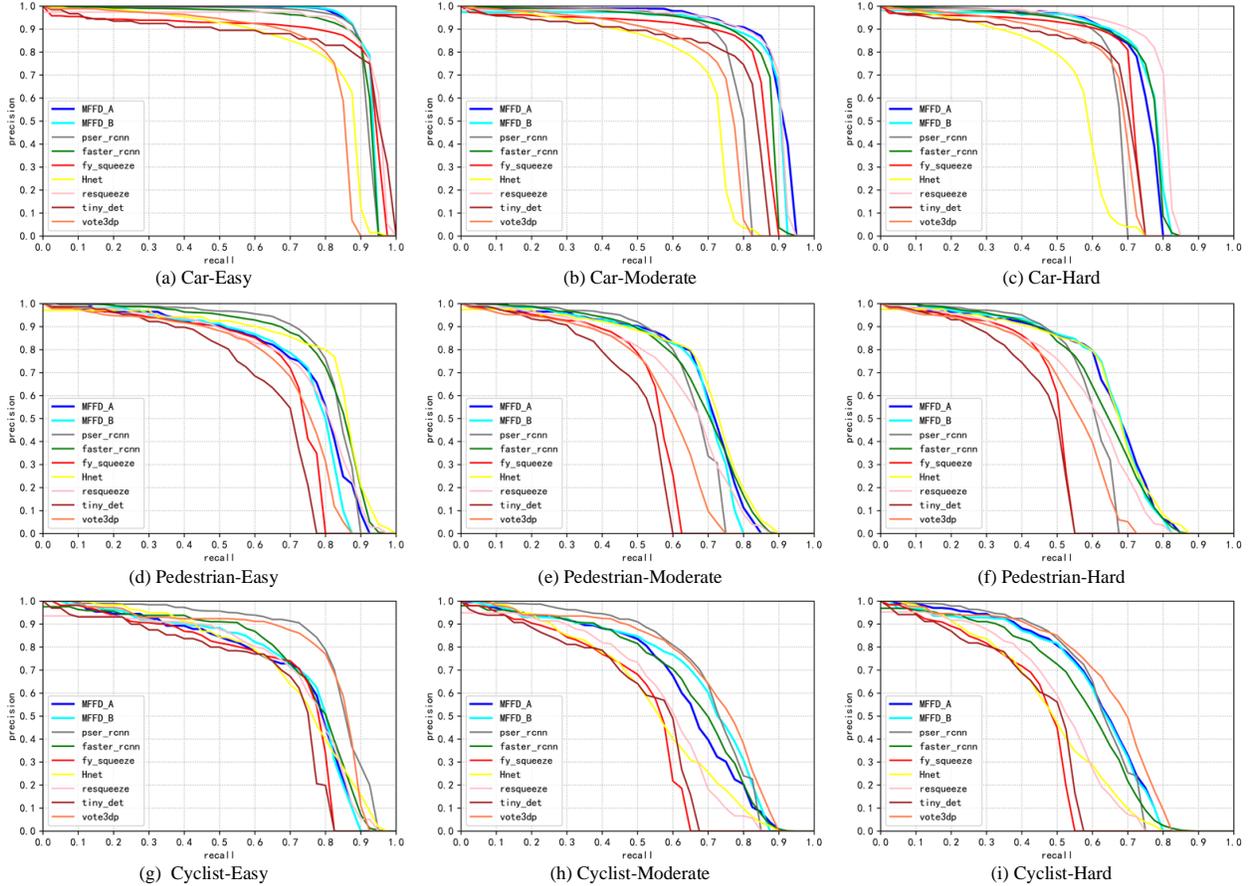

Fig. 6. The Precision-Recall curves for Car, Pedestrian and Cyclist on different difficulty levels. The proposed MFFD are drawn in blue lines.

*D. Model size and running time*

To demonstrate the computational efficiency of the proposed detection model, Table 3 provides the results in terms of the average test time (seconds) per image. We evaluate the testing time on different platforms, such as CPU, GPU and embedded GPUs. Specifically, the CPU is i7-6820HQ with 2.70 GHz, the GPU is Nvidia 1080Ti, and the embedded GPUs are Nvidia Drive PX2 and Jetson TX2.

Different deep learning backend may lead to different running time. In order to make a fair comparison, all the baselines in this subsection have exactly the same backend and running environment, therefore, their model size and inference time can be compared directly. Specifically, since our backend is Darknet, we adopt two widely used lightweight and fast network models YOLOv2-Tiny [33], YOLOv3-Tiny [58] as the baselines and compare them with the proposed MFFD_A and MFFD_B detectors. The model size, mAP and running time of all the methods are shown in Table 3. It should be noted that even for the baselines YOLOv2-Tiny [33] and YOLOv3-Tiny [58], we have retrained the model on the KITTI dataset, and using the same image size and training parameters as MFFD.

As shown in Table 3, we have following observations:
1) From accuracy point of view, the proposed methods outperform the baselines by a big margin. Specifically, MFFD_B's mAP is 72.71%, which is more than 10% higher than YOLOv3-Tiny. MFFD_A is the runner-up with 68.48% mAP.
2) In terms of model size, MFFD_A is the best one with only 29.7 MB, YOLOv3-Tiny is the runner-up with 34.7 MB,

MFFD_B and YOLOv2-Tiny are relatively larger with 55.6MB and 63.5MB respectively.

3) Regarding running time, both YOLOv2-Tiny and YOLOv3-Tiny are faster than the proposed methods on all the platforms. But it should be noted that, even the speeds are lower, but the differences are not substantial, and actually all the methods are quite fast. Take the slowest one MFFD_B for example, it still can run with 100 fps on the embedded GPU Jetson TX2.

Table 2. Comparison of different methods on the KITTI validation set.

| Method | Car | | | Pedestrian | | | Cyclist | | |
|---|---|---|---|---|---|---|---|---|---|
| | Easy | Mod | Hard | Easy | Mod | Hard | Easy | Mod | Hard |
| Pose-RCNN [52] | 88.89 | 75.74 | 61.86 | **77.79** | 63.38 | 57.42 | **80.16** | 62.25 | 55.09 |
| Faster-RCNN [12] | 87.90 | 79.11 | 70.19 | **78.35** | **65.91** | 61.19 | 71.41 | 62.81 | 55.44 |
| FYSqueeze [53] | 84.06 | 76.73 | 67.96 | 66.07 | 52.60 | 48.40 | 67.03 | 48.80 | 43.82 |
| HNet [54] | 77.09 | 66.00 | 53.89 | 77.39 | **66.74** | 62.26 | 69.71 | 54.10 | 48.02 |
| ReSqueeze [56] | 87.12 | **85.74** | **77.02** | 72.78 | 61.25 | 57.43 | 68.34 | 54.93 | 49.19 |
| tiny-det [55] | 81.88 | 73.46 | 63.70 | 62.02 | 47.81 | 45.53 | 63.78 | 50.48 | 44.23 |
| Vote3Deep [57] | 76.95 | 68.39 | 63.22 | 67.94 | 55.38 | 52.62 | **76.49** | **67.96** | **62.88** |
| MFFD_A (Proposed) | **91.06** | **86.48** | 71.62 | 74.56 | 65.46 | **63.86** | 68.96 | 62.69 | **58.48** |
| MFFD_B (Proposed) | **91.16** | 84.01 | **72.43** | 73.08 | 64.32 | **62.96** | 71.55 | **65.62** | 58.21 |

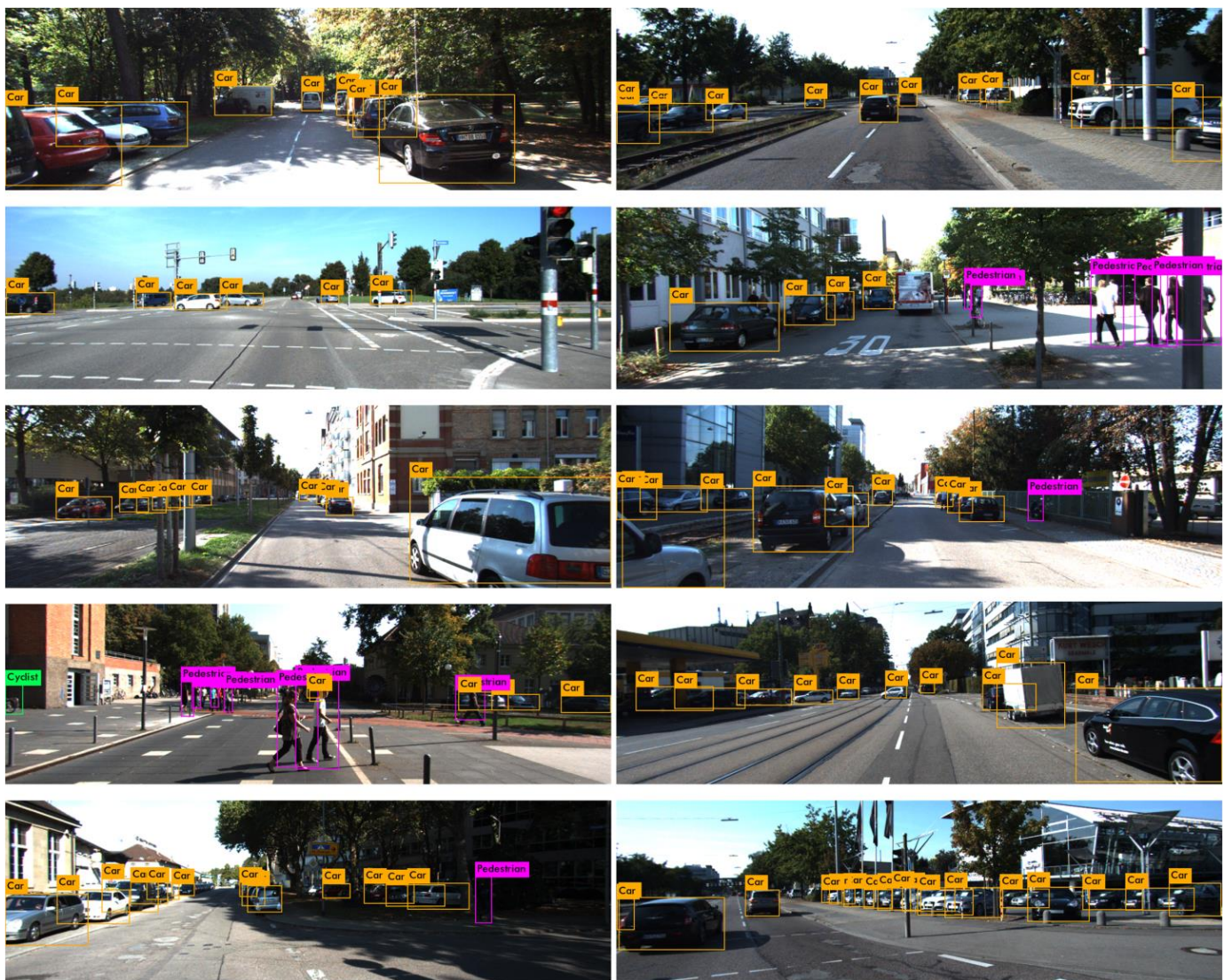

Fig. 7. Selected examples of object detection results on KITTI testing set using the MFFD_A. For each image, one color corresponds to one object category in that image.

Table 3. Comparison of different models in terms of mAP, size and inference time. (IOU=0.5)

| Model | mAP | Size (MB) | GPU time 1080Ti (s) | Embed GPU Drive PX2 (s) | Embed GPU Jetson TX2 (s) | CPU time (s) |
|---|---|---|---|---|---|---|
| YOLOv2-Tiny [33] | 50.62 | 63.5 | **0.0037** | **0.0037** | **0.0035** | **0.6975** |
| YOLOv3-Tiny [58] | 61.93 | **34.7** | **0.0032** | **0.0047** | **0.0067** | **0.6098** |
| MFFD_A | **68.48** | 29.7 | 0.0044 | 0.0060 | 0.0093 | 0.9729 |
| MFFD_B | **72.71** | 55.6 | 0.0055 | 0.0062 | 0.0098 | 1.0630 |

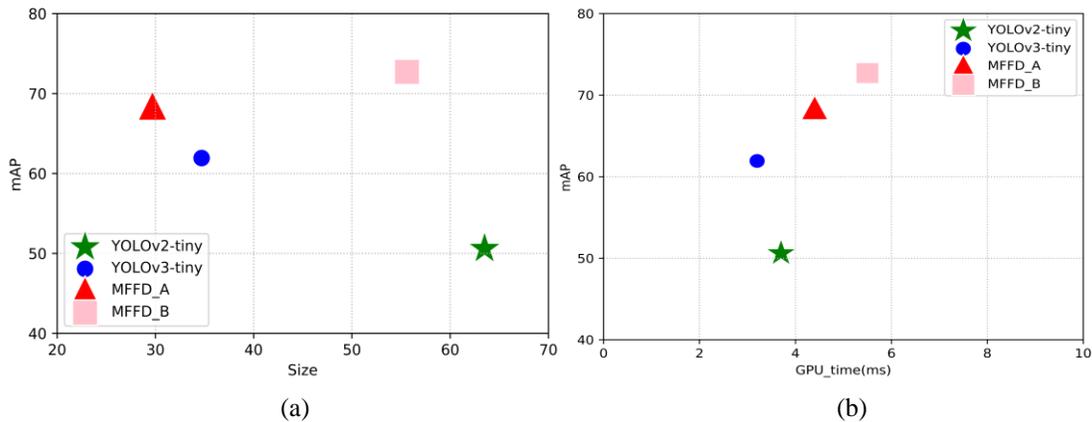

Fig. 8. (a): Size versus Accuracy(mAP). (b): Speed versus accuracy. If the method has smaller size (or less time) while getting higher mAP, it is better than other methods. So the methods in the upper-left of the plots represent better results.

In addition, we plot the model size VS. mAP and GPU_time VS. mAP figures in Fig 8 (a) and (b). (a) shows that MFFD_A achieves better performance with smaller size and getting higher mAP. (b) shows that there is a trade-off between precision and speed. So, generally, if the accuracy, model size and inference time are considered jointly, MFFD_A is a very competitive model; otherwise, MFFD_B is good choice in terms of performance and testing speed.

## V. Conclusion

In this paper, we have designed a new light-weighted network, modular feature fusion detector (MFFD), for detecting the objects on the road in the application background of ADAS. Firstly, components of the network are not individual layers but the carefully designed modules, and building the network from these modules ensures high performance and computation efficiency. Secondly, two different feature fusion models have been developed to add contextual information for detection. Benefited from lightweight modularized design, MFFD can be deployed on the platforms with limited memory and computation resources. The proposed methods have been tested on different platforms, such as CPU, GPU and embedded GPUs. Experimental evaluations on the public available KITTI detection datasets demonstrated that the presented methods are very competitive in terms of performance, model size and inference time. Even the methods are proposed in the background of ADAS, they may have potential application in the other embedded systems, such as robot or drone.

References

[1] K. He, X. Zhang, S. Ren, and J. Sun, "Deep Residual Learning for Image Recognition," in Computer Vision and Pattern Recognition, 2016, pp. 770-778.

[2] G. Huang, Z. Liu, K. Q. Weinberger, and L. V. Der Maaten, "Densely Connected Convolutional Networks," in Computer Vision and Pattern Recognition, 2017.

[3] K. Simonyan, and A. Zisserman, "Very Deep Convolutional Networks for Large-Scale Image Recognition," in International Conference on Learning Representations, 2015.

[4] C. Szegedy, W. Liu, Y. Jia, P. Sermanet, S. E. Reed, D. Anguelov, D. Erhan, V. Vanhoucke, and A. Rabinovich, "Going Deeper with Convolutions," in IEEE Conference on Computer Vision and Pattern Recognition 2015, pp. 1-9.

[5] A. Krizhevsky, I. Sutskever, and G. E. Hinton, "ImageNet Classification with Deep Convolutional Neural Networks," in Neural Information Processing Systems, 2012, pp. 1097-1105.

[6] A. Kamel, B. Sheng, P. Yang, P. Li, R. Shen, and D. D. Feng, "Deep Convolutional Neural Networks for Human Action Recognition Using Depth Maps and Postures," *IEEE Transactions on Systems, Man, and Cybernetics: Systems* vol. pp, no. 99, 2018.

[7] N. Sun, W. Li, J. Liu, G. Han, and C. Wu, "Fusing Object Semantics and Deep Appearance Features for Scene Recognition," *IEEE Transactions on Circuits and Systems for Video Technology*, vol. PP, no. 99, 2018.

[8] J. Dai, Y. Li, K. He, and J. Sun, "R-FCN: Object Detection via Region-based Fully Convolutional Networks," in Neural Information Processing Systems, 2016, pp. 379-387.

[9] R. B. Girshick, J. Donahue, T. Darrell, and J. Malik, "Rich Feature Hierarchies for Accurate Object Detection and Semantic Segmentation," in Computer Vision and Pattern Recognition, 2014, pp. 580-587.

[10] W. Liu, D. Anguelov, D. Erhan, C. Szegedy, S. E. Reed, C. Fu, and A. C. Berg, "SSD: Single Shot MultiBox Detector," in European Conference on Computer Vision, 2016, pp. 21-37.

> REPLACE THIS LINE WITH YOUR PAPER IDENTIFICATION NUMBER (DOUBLE-CLICK HERE TO EDIT) <       11
[11] J. Redmon, S. K. Divvala, R. B. Girshick, and A. Farhadi, "You Only Look Once: Unified, Real-Time Object Detection," in Computer Vision and Pattern Recognition, 2016, pp. 779-788.

[12] S. Ren, K. He, R. B. Girshick, and J. Sun, "Faster R-CNN: Towards Real-Time Object Detection with Region Proposal Networks," *IEEE Transactions on Pattern Analysis and Machine Intelligence,* vol. 39, no. 6, pp. 1137-1149, 2017.

[13] Z. Shen, Z. Liu, J. Li, Y. G. Jiang, Y. Chen, and X. Xue, "DSOD: Learning Deeply Supervised Object Detectors from Scratch," in International Conference on Computer Vision, 2017, pp. 1937-1945.

[14] K. Muhammad, J. Ahmad, Z. Lv, P. Bellavista, P. Yang, and S. W. Baik, "Efficient Deep CNN-Based Fire Detection and Localization in Video Surveillance Applications," *IEEE Transactions on Systems, Man, and Cybernetics: Systems,* vol. PP, no. 99, 2018.

[15] S. Bąk, M. S. Biagio, R. Kumar, V. Murino, and F. Brémond, "Exploiting Feature Correlations by Brownian Statistics for People Detection and Recognition," *IEEE Transactions on Systems, Man, and Cybernetics: Systems,* vol. 47, no. 9, pp. 2538 - 2549, 2017.

[16] S. López-Tapia, R. Molina, and N. P. d. l. Blanca, "Deep CNNs for Object Detection using Passive Millimeter Sensors," *IEEE Transactions on Circuits and Systems for Video Technology,* vol. pp, no. 99, 2018.

[17] L. Chen, G. Papandreou, I. Kokkinos, K. Murphy, and A. L. Yuille, "Semantic Image Segmentation with Deep Convolutional Nets and Fully Connected CRFs," in International Conference on Learning Representations, 2015.

[18] B. Hariharan, P. Arbelaez, R. B. Girshick, and J. Malik, "Hypercolumns for object segmentation and fine-grained localization," in Computer Vision and Pattern Recognition, 2015, pp. 447-456.

[19] J. Long, E. Shelhamer, and T. Darrell, "Fully convolutional networks for semantic segmentation," in Computer Vision and Pattern Recognition, 2015, pp. 3431-3440.

[20] F. Yu, and V. Koltun, "Multi-Scale Context Aggregation by Dilated Convolutions," in International Conference on Learning Representations, 2016.

[21] O. Russakovsky, J. Deng, H. Su, J. Krause, S. Satheesh, S. Ma, Z. Huang, A. Karpathy, A. Khosla, and M. S. Bernstein, "ImageNet Large Scale Visual Recognition Challenge," *International Journal of Computer Vision,* vol. 115, no. 3, pp. 211-252, 2015.

[22] R. B. Girshick, J. Donahue, T. Darrell, and J. Malik, "Rich Feature Hierarchies for Accurate Object Detection and Semantic Segmentation," in IEEE Conference on Computer Vision and Pattern Recognition, 2014, pp. 580-587.

[23] J. Redmon, S. K. Divvala, R. B. Girshick, and A. Farhadi, "You Only Look Once: Unified, Real-Time Object Detection," in IEEE Conference on Computer Vision and Pattern Recognition, 2016, pp. 779-788.

[24] K. He, X. Zhang, S. Ren, and J. Sun, "Deep Residual Learning for Image Recognition," in IEEE Conference on Computer Vision and Pattern Recognition, 2016, pp. 770-778.

[25] J. R. R. Uijlings, K. E. A. V. De Sande, T. Gevers, and A. W. M. Smeulders, "Selective Search for Object Recognition," *International Journal of Computer Vision,* vol. 104, no. 2, pp. 154-171, 2013.

[26] C. L. Zitnick, and P. Dollar, "Edge Boxes: Locating Object Proposals from Edges." pp. 391-405.

[27] J. Redmon, and A. Farhadi, "YOLO9000: Better, Faster, Stronger," in IEEE Conference on Computer Vision and Pattern Recognition, 2017.

[28] M. Everingham, L. Van Gool, C. K. I. Williams, J. M. Winn, and A. Zisserman, "The Pascal Visual Object Classes (VOC) Challenge," *International Journal of Computer Vision,* vol. 88, no. 2, pp. 303-338, 2010.

[29] A. Geiger, P. Lenz, and R. Urtasun, "Are we ready for autonomous driving? The KITTI vision benchmark suite," in IEEE Conference on Computer Vision and Pattern Recognition 2012, pp. 3354-3361.

[30] J. Long, E. Shelhamer, and T. Darrell, "Fully convolutional networks for semantic segmentation," in IEEE Conference on Computer Vision and Pattern Recognition, 2015, pp. 3431-3440.

[31] J. Redmon, and A. Farhadi, "YOLOv3: An Incremental Improvement," *arXiv:1804.02767*, 2018.

[32] R. B. Girshick, "Fast R-CNN," in International Conference on Computer Vision, 2015, pp. 1440-1448.

[33] J. Redmon, and A. Farhadi, "YOLO9000: Better, Faster, Stronger," in Computer Vision and Pattern Recognition, 2017.

[34] C. Szegedy, S. Ioffe, V. Vanhoucke, and A. A. Alemi, "Inception-v4, Inception-ResNet and the Impact of Residual Connections on Learning," in National Conference on Artificial Intelligence, 2016, pp. 4278-4284.

[35] G. Huang, Z. Liu, K. Q. Weinberger, and L. V. Der Maaten, "Densely Connected Convolutional Networks," in IEEE Conference on Computer Vision and Pattern Recognition, 2017.

[36] S. Han, H. Mao, and W. J. Dally, "Deep Compression: Compressing Deep Neural Networks with Pruning, Trained Quantization and Huffman Coding," in International Conference on Learning Representations, 2016.

[37] F. N. Iandola, S. Han, M. W. Moskewicz, K. Ashraf, W. J. Dally, and K. Keutzer, "SqueezeNet: AlexNet-level accuracy with 50x fewer parameters and <0.5MB model size," in International Conference on Learning Representations, 2017.

[38] X. Zhang, X. Zhou, M. Lin, and J. Sun, "ShuffleNet: An Extremely Efficient Convolutional Neural Network for Mobile Devices," in IEEE Conference on Computer Vision and Pattern Recognition, 2017.

[39] M. Sandler, A. Howard, M. Zhu, A. Zhmoginov, and L.-C. Chen, "Inverted Residuals and Linear Bottlenecks: Mobile Networks forClassification, Detection and Segmentation," *arXiv:1801.04381*, 2018.

[40] S. Bell, C. L. Zitnick, K. Bala, and R. B. Girshick, "Inside-Outside Net: Detecting Objects in Context with Skip Pooling and Recurrent Neural Networks," in IEEE Conference on Computer Vision and Pattern Recognition, 2016, pp. 2874-2883.

[41] D. Park, D. Ramanan, and C. C. Fowlkes, "Multiresolution models for object detection," in European Conference on Computer Vision, 2010, pp. 241-254.

[42] C. Zhu, Y. Zheng, K. Luu, and M. Savvides, "CMS-RCNN: Contextual Multi-Scale Region-based CNN for Unconstrained Face Detection," *arXiv:1606.05413*, 2016.

[43] J. Ren, X. Chen, J. Liu, W. Sun, J. Pang, Q. Yan, Y. Tai, and L. Xu, "Accurate Single Stage Detector Using Recurrent Rolling Convolution," in IEEE Conference on Computer Vision and Pattern Recognition, 2017, pp. 752-760.

[44] C. Szegedy, V. Vanhoucke, S. Ioffe, J. Shlens, and Z. Wojna, "Rethinking the Inception Architecture for Computer Vision," in Computer Vision and Pattern Recognition, 2016, pp. 2818-2826.

[45] K. He, X. Zhang, S. Ren, and J. Sun, "Delving Deep into Rectifiers: Surpassing Human-Level Performance on ImageNet Classification," in International Conference on Computer Vision, 2015, pp. 1026-1034.

[46] R. K. Srivastava, K. Greff, and J. Schmidhuber, "Training very deep networks," in Neural Information Processing Systems, 2015, pp. 2377-2385.

[47] Z. Liu, J. Li, Z. Shen, G. Huang, S. Yan, and C. Zhang, "Learning Efficient Convolutional Networks through Network Slimming," in International Conference on Computer Vision, 2017.

[48] M. Denil, B. Shakibi, L. Dinh, M. Ranzato, and N. De Freitas, "Predicting Parameters in Deep Learning," in Neural Information Processing Systems, 2013, pp. 2148-2156.

[49] V. Nair, and G. E. Hinton, "Rectified Linear Units Improve Restricted Boltzmann Machines," in International Conference on Machine Learning, 2010, pp. 807-814.

[50] A. Geiger, P. Lenz, and R. Urtasun, "Are we ready for autonomous driving? The KITTI vision benchmark suite," in Computer Vision and Pattern Recognition, 2012, pp. 3354-3361.

[51] X. Chen, K. Kundu, Y. Zhu, A. Berneshawi, H. Ma, S. Fidler, and R. Urtasun, "3D object proposals for accurate object class detection," in Neural Information Processing Systems, 2015.

[52] M. Braun, Q. Rao, and Y. Wang, "Pose-RCNN: Joint object detection and pose estimation using 3D object proposals," in Intelligent Transportation Systems, 2016.

[53] "FYSqueeze," http://www.cvlibs.net/datasets/kitti/eval_object.php.

[54] "HNet," http://www.cvlibs.net/datasets/kitti/eval_object_detail.php?&result=1910b3cb76f69c31abfa242d1dd519ae387d26d1.

[55] "tiny-det," http://www.cvlibs.net/datasets/kitti/eval_object_detail.php?&result=e38d125dda923331a4e090a50f8728a15b0712bb.

[56] "ReSqueeze," http://www.cvlibs.net/datasets/kitti/eval_object_detail.php?&result=f47804342b172f8a110dcbba0e690364dd6996a8.

[57] M. Engelcke, D. Rao, D. Z. Wang, C. H. Tong, and I. Posner, "Vote3Deep: Fast Object Detection in 3D Point Clouds Using Efficient Convolutional Neural Networks," in International Conference on Robotics and Automation, 2017.

[58] J. Redmon, and A. Farhadi, *YOLOv3: An Incremental Improvement*, arXiv:1804.02767, 2018.